\documentclass{article} 
\usepackage{iclr2027_conference,times}

\usepackage{amsmath,amsfonts,bm}

\def\eqref#1{equation~\ref{#1}}

\def\1{\bm{1}}

\DeclareMathAlphabet{\mathsfit}{\encodingdefault}{\sfdefault}{m}{sl}
\SetMathAlphabet{\mathsfit}{bold}{\encodingdefault}{\sfdefault}{bx}{n}

\usepackage{hyperref}
\usepackage{url}

\usepackage{graphicx}
\usepackage{booktabs}       
\usepackage{amsfonts}       
\usepackage{nicefrac}       
\usepackage{xcolor}         

\usepackage{amsmath}
\usepackage{amssymb}
\usepackage{graphicx}
\usepackage{multirow}
\usepackage{bm}
\usepackage{array}
\usepackage{makecell}
\usepackage{adjustbox}
\usepackage[table]{xcolor}
\usepackage{pifont}
\usepackage{algorithm}
\usepackage{algpseudocode}

\usepackage{cuted}
\usepackage{wrapfig}
\usepackage{caption}

\usepackage{booktabs}

\newcommand{\sd}[1]{\ensuremath{\,\pm\,#1}}
\newcommand{\method}{SAREO-FM}
\newcommand{\sarmaestar}{SARMAE$^\ast$}
\newcommand{\optionalinput}[1]{\IfFileExists{#1}{\input{#1}}{}}

\title{SAREO-FM: Decoupled Semantic Supervision\\for SAR-EO Foundation Models}

\author{
Jeonghyeok Do \quad Munchurl Kim \thanks{Corresponding author.}\\
Korea Advanced Institute of Science and Technology (KAIST)\\
\texttt{\{ehwjdgur0913,mkimee\}@kaist.ac.kr}\\[4pt]
{\small Project Page: \url{https://kaist-viclab.github.io/SAREO-FM_site/}}
}

\iclrfinalcopy
\begin{document}

\maketitle

\vspace{-0.4cm}
\begin{abstract}
Synthetic aperture radar (SAR) and electro-optical (EO) imagery provide complementary observations: SAR enables day-and-night, weather-resilient sensing, whereas EO provides rich appearance and fine-grained semantic cues. We introduce \textbf{SAREO-FM}, which avoids forcing a single token stream to serve two distinct roles: modality tokens preserve how each sensor observes the scene through masked reconstruction, while learnable semantic queries capture what the scene contains under guidance from a pretrained vision foundation model (VFM). By jointly encoding these queries with SAR and EO tokens, the queries acquire modality-grounded semantic context, while the modality-token outputs remain the explicit targets of masked reconstruction. This design assigns semantic and reconstruction supervision to separate token streams while preserving their interaction within the shared encoder. Pretrained on the million-scale SAR-1M corpus, SAREO-FM achieves strong unimodal transfer for both SAR-only and EO-only inputs, while delivering substantial gains from joint SAR--EO observations on tasks that benefit from complementary sensing.
\end{abstract}    
\vspace{-0.2cm}

\section{Introduction}
\label{sec:intro}
\begin{wrapfigure}{r}{0.48\textwidth}
  \centering
  \vspace{-0.5cm}
  \includegraphics[width=0.48\textwidth]{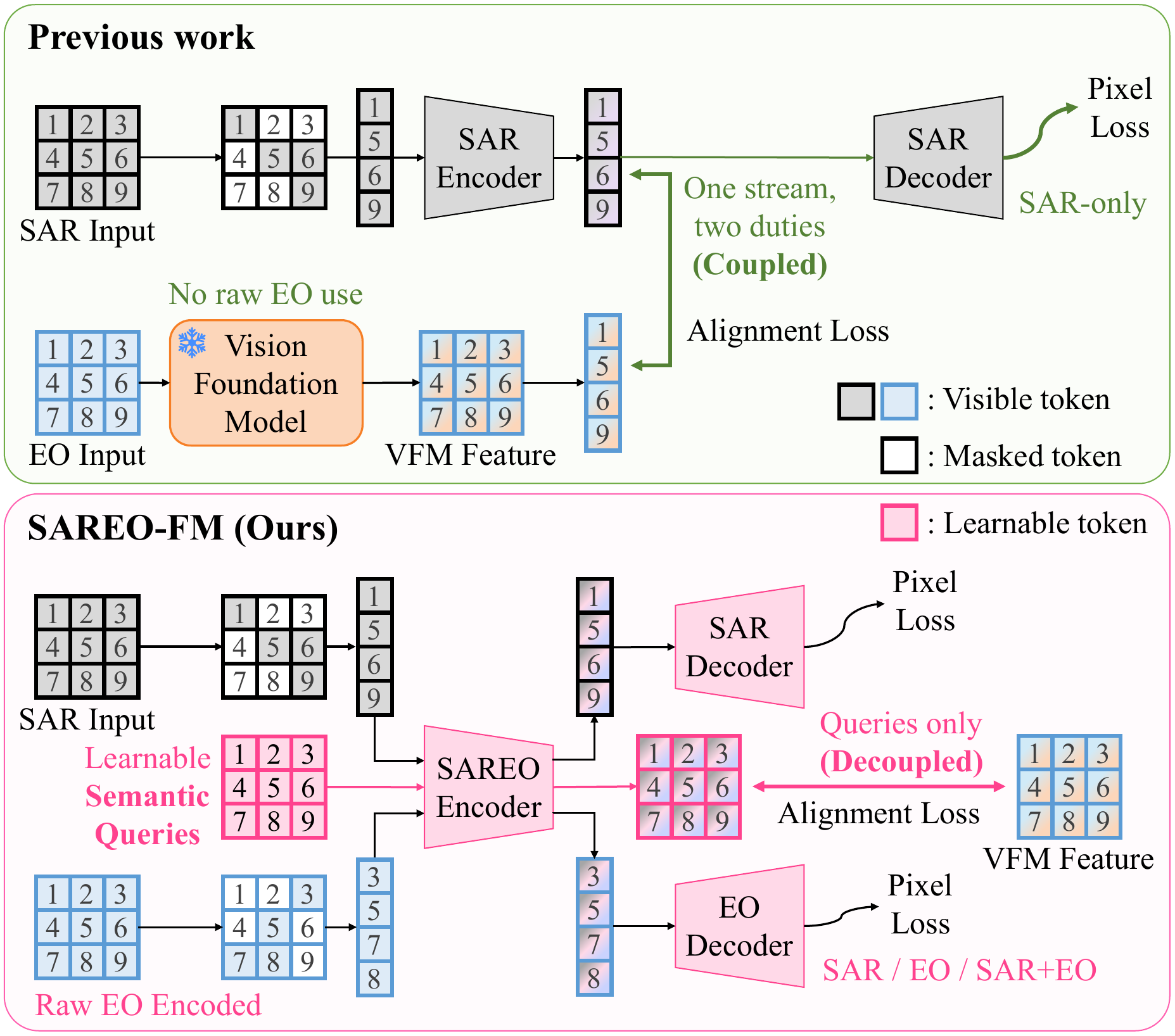}
  \vspace{-0.55cm}
  \caption{\textbf{Motivation of SAREO-FM.}
  Prior methods directly align image tokens with EO-derived VFM features, coupling semantic alignment with pixel reconstruction. SAREO-FM instead supervises learnable semantic queries while reconstructing SAR and EO from modality tokens, separating the destinations of the two objectives within a unified encoder.}
  \label{fig:motivation}
  \vspace{-0.5cm}
\end{wrapfigure}
Synthetic aperture radar (SAR) and electro-optical (EO) imagery provide complementary views of the Earth. SAR measures microwave backscatter and enables day-and-night, weather-resilient observation, whereas EO imagery captures rich spectral appearance, texture, and fine-grained semantic cues. Their complementary information benefits numerous applications, including land-cover mapping, object detection, change and damage assessment, cross-modal retrieval, cloud removal, and SAR-to-EO translation~\cite{schmitt2016datafusion,zhu2017deeprs,do2026c}. So, a practical SAR--EO foundation model (FM) should learn strong representations from either modality alone while exploiting their complementary evidence when paired observations are available.

Recent geospatial foundation models (GFMs) \cite{cong2022satmae,copernicusfm,mars} have demonstrated the value of pretraining across multiple sensors and data sources. However, simply combining modalities does not resolve a fundamental representation-learning challenge. SAR and EO capture the same geographic scene through substantially different sensing mechanisms and visual statistics. Their representations must preserve modality-specific details while also learning high-level semantics that transfer across modalities and downstream tasks.

Masked autoencoders (MAEs)~\cite{he2022mae} provide a natural framework for preserving sensing-specific information through masked image reconstruction. Recent approaches further enhance MAE representations using semantic guidance from pretrained vision foundation models (VFMs)~\cite{yu2024representation,yao2025vavae,liu2026sarmae}. As illustrated in Figure~\ref{fig:motivation}(a), prior representative methods directly align SAR or EO image tokens with VFM features. Consequently, the same image tokens are explicitly optimized both to preserve modality-specific information for pixel reconstruction and to match high-level EO-derived semantics. To our best knowledge, we \textit{firstly} address that this coupling raises a central question: \emph{where should semantic supervision be applied so that shared semantic learning does not directly constrain modality-specific reconstruction?}

We address this question with \textbf{SAREO-FM}, a unified MAE-based FM with \emph{decoupled semantic supervision}. As shown in Figure~\ref{fig:motivation}(b), our SAREO-FM processes SAR tokens, EO tokens, and learnable semantic queries using a shared and unified encoder. The semantic queries interact with all available image tokens, and are aligned patch-wise with features from a frozen pretrained VFM. In parallel, the modality-token outputs are passed to separate SAR and EO decoders for masked pixel reconstruction. Importantly, the semantic queries are excluded from the reconstruction decoders. Thus, the SAREO-FM separates the explicit destinations of semantic alignment and pixel reconstruction while retaining information exchange among all three streams within the shared and unified encoder. To support different input configurations, we further introduce a mixed modality masking strategy that combines independent masking, spatially shared masking, and complete dropping of either modality. This exposes the same encoder to joint SAR--EO, SAR-only, and EO-only observations during pretraining. At downstream transfer, semantic queries always participate in encoding, while task heads can selectively consume modality-token outputs, semantic-query outputs, or both according to specific tasks.
We pretrain SAREO-FM on SAR--EO observations from the million-scale SAR-1M corpus~\cite{liu2026sarmae}, treating both SAR and EO imagery as raw learning inputs rather than using EO solely to obtain teacher features. 
We evaluate the resulting representations on image classification and multi-label recognition, and further examine qualitative transfer to
dense prediction under SAR-only, EO-only, and joint SAR--EO settings. Our SAREO-FM achieves strong unimodal transfer and substantial task-dependent improvements from joint observations. Extensive ablations further validate the effects of semantic-query supervision and its separation from reconstruction.

Our contributions are as follows:

\begin{itemize}
    \item We introduce \textbf{SAREO-FM}, a unified MAE-based foundation model that learns from raw SAR and EO imagery using a shared encoder and supports SAR-only, EO-only, and joint SAR--EO transfer.

    \item We propose \emph{decoupled semantic supervision}, which assigns EO-VFM alignment to learnable semantic queries and modality-specific reconstruction to image tokens, while allowing the three streams to interact within the shared encoder.

    \item We develop a mixed modality masking strategy that exposes the encoder to paired and missing-modality inputs. Comprehensive evaluations demonstrate strong unimodal transfer and substantial task-dependent gains from complementary SAR--EO observations.
\end{itemize}

\section{Related Work}
\label{sec:related}

\subsection{Masked Multimodal Modeling}
MAE learns visual representations by reconstructing heavily masked image patches with an asymmetric encoder--decoder~\cite{he2022mae}. Subsequent work extends masked modeling beyond RGB imagery. MultiMAE~\cite{bachmann2022multimae} jointly models RGB, depth, and semantic maps, while 4M~\cite{mizrahi2023fourm} and 4M-21~\cite{bachmann2024fourm21} scale masked modeling to a broad collection of input and output modalities, including images, geometry, semantics, and feature representations. These approaches demonstrate that masked modeling can effectively integrate heterogeneous signals. Our SAREO-FM follows this general framework but introduces separate token streams as the explicit destinations of pixel-reconstruction and frozen-VFM supervision.

\subsection{Geospatial Foundation Models (GFMs)}
Self-supervised SAR representation learning has progressed from predictive and reconstruction-based approaches such as SAR-JEPA~\cite{sarjepa} to large-scale FMs such as SARATR-X~\cite{saratrx} and SARMAE~\cite{liu2026sarmae}. In parallel, multimodal Earth-observation pretraining has explored masked reconstruction, cross-modal alignment, and sensor-aware architectures. CROMA~\cite{croma} combines contrastive radar--optical learning with masked autoencoding, while SatMAE~\cite{cong2022satmae} and SatMAE++~\cite{noman2024rethinking} address multispectral and multitemporal observations. Recent works such as MMEarth~\cite{nedungadi2024mmearth}, msGFM~\cite{han2024msgfm}, Copernicus-FM~\cite{copernicusfm}, and MaRS~\cite{mars} further broaden the range of sensors, resolutions, and downstream tasks. DeCUR~\cite{wang2024decur} instead partitions modality-common and modality-specific embedding dimensions through redundancy reduction. Whereas these methods address sensor heterogeneity through alignment, representation partitioning, or architectural adaptation, we focus on separating the explicit supervisions assigned to \textit{decoupled semantic and reconstruction representations} of SAR and EO modalities in a shared latent space by a unified encoder using \textit{learnable semantic queries}.

\subsection{Semantic Feature Alignment}
Aligning learned representations with strong pretrained encoders has proven effective across representation and generative learning, as exemplified by REPA~\cite{yu2024representation} and VA-VAE~\cite{yao2025vavae}. In SAR pretraining, SARMAE~\cite{liu2026sarmae} introduces semantic guidance from a frozen DINOv3~\cite{simeoni2025dinov3} teacher by directly aligning SAR encoder tokens with features extracted from paired EO imagery. The EO image therefore provides teacher-side supervision but is not encoded as an input modality by the student. CoDe-MAE~\cite{codemae}, the closest concurrent multimodal MAE, jointly handles SAR and EO through optical-token distillation, conditioned on (i) contrastive alignment and (ii) degraded cross-modal reconstruction. Our SAREO-FM differs in the destination of semantic supervision: frozen-VFM features supervise only the persistent semantic-query outputs, while SAR and EO token outputs are used for modality-specific reconstruction.
\section{SAREO-FM}
\label{sec:method}

\begin{figure*}[t]
  \centering
  \includegraphics[width=0.9\textwidth]{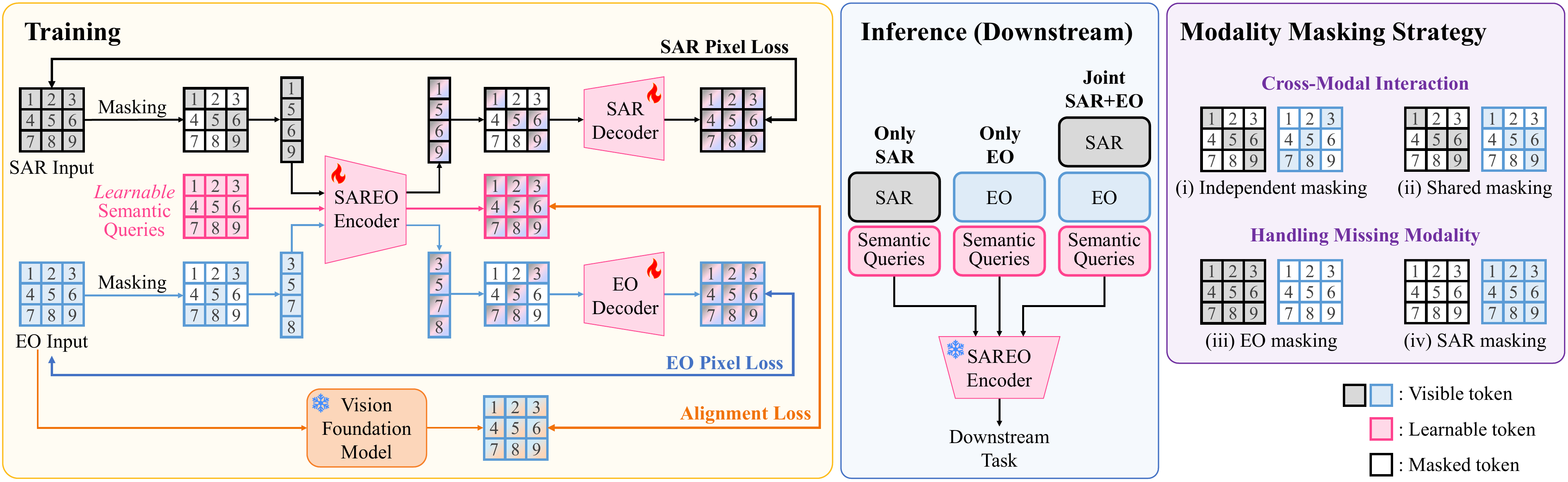}
  \caption{\textbf{Overview of SAREO-FM.} Visible SAR and EO tokens interact with learnable semantic queries through a unified encoder. The modality-token outputs are decoded independently for masked pixel reconstruction, whereas the semantic-query outputs are aligned patch-wise with features from a frozen pretrained VFM. Semantic queries are excluded from the reconstruction decoders, separating the destinations of pixel-level and semantic supervision. Modality masking exposes the encoder to SAR-only, EO-only, and joint SAR--EO inputs; during downstream transfer, task heads selectively consume modality-token outputs, semantic-query outputs, or both.}
  \label{fig:overview}
\end{figure*}

\subsection{Overview}
\label{sec:overview}

Our SAREO-FM learns modality-specific and cross-modal representations from SAR and EO imagery within a unified masked-autoencoding framework. Figure~\ref{fig:overview} illustrates the overall architecture of SAREO-FM. Our key design principle is to separate the destinations of pixel-level reconstruction and high-level semantic supervision. To this end, the SAREO-FM operates on three token streams: SAR image tokens, EO image tokens, and learnable semantic queries, which interact through a shared SAREO Encoder.
The semantic queries aggregate cross-modal context from the visible SAR and EO tokens and are aligned with patch-level features extracted by a frozen pretrained VFM. In parallel, the encoded SAR and EO tokens are passed to modality-specific decoders for masked image reconstruction, thereby preserving sensing-specific appearance and structure. The semantic queries are excluded from the decoding stage, preventing the reconstruction decoders from directly relying on the semantically supervised stream. After pretraining, the VFM teacher and reconstruction decoders are discarded, and the unified encoder serves as our SAREO-FM for SAR, EO, or joint SAR--EO observations.

\subsection{Unified Three-Stream SAREO Encoder}
\label{sec:encoder}

Let $\mathbf{X}_{s}\in\mathbb{R}^{H\times W\times C_s}$ and
$\mathbf{X}_{e}\in\mathbb{R}^{H\times W\times C_e}$ denote a spatially registered SAR--EO pair, where $s$ and $e$ index the SAR and EO modalities, respectively. Each image is divided into $N$ non-overlapping patches. For modality $m\in\{s,e\}$, the $i$-th flattened patch
$\mathbf{x}_{p,m}^{i}$ is projected into a shared $D$-dimensional embedding space using a modality-specific linear projection $\mathbf{E}_m$. The resulting image token is
\begin{equation}
\mathbf{y}_{m}^{i}
=
\mathbf{x}_{p,m}^{i}\mathbf{E}_m
+
\mathbf{p}^{i}
+
\mathbf{a}_{m}^{\mathrm{enc}},
\qquad
i=1,\ldots,N,
\label{eq:image_token}
\end{equation}
where $\mathbf{p}^{i}$ is a shared 2D sine--cosine positional embedding and
$\mathbf{a}_{m}^{\mathrm{enc}}$ is a learnable modality embedding for the encoder. We denote the full token sequence by
$\mathbf{Y}_m=[\mathbf{y}_m^1,\ldots,\mathbf{y}_m^N]$. For each modality, we sample a binary mask
$\mathcal{M}_m\in\{0,1\}^{N}$, where
$\mathcal{M}_m[i]=1$ or $0$ indicates a masked or visible patch. We define the masked and visible index sets as
\begin{equation}
\Omega_m
=
\{i\mid\mathcal{M}_m[i]=1\},
\quad
\mathcal{V}_m
=
\{1,\ldots,N\}\setminus\Omega_m.
\label{eq:index_sets}
\end{equation}
Following masked autoencoders, only the visible image tokens are provided to the encoder:
\begin{equation}
\mathbf{Y}_m^{\mathrm{vis}}
=
[\mathbf{y}_m^i]_{i\in\mathcal{V}_m}.
\label{eq:visible_tokens}
\end{equation}

\noindent In addition to the image streams, we introduce a spatially organized set of $N$ learnable semantic queries,
$\mathbf{Q}=\{\mathbf{q}^{i}\}_{i=1}^{N}$. Each query corresponds to one image-patch location and is augmented with the same positional embedding used for the image tokens, together with a dedicated query-stream embedding:
\begin{equation}
\widetilde{\mathbf{q}}^{i}
=
\mathbf{q}^{i}
+
\mathbf{p}^{i}
+
\mathbf{a}_{q}^{\mathrm{enc}},
\qquad
\widetilde{\mathbf{Q}}
=
[\widetilde{\mathbf{q}}^{1},\ldots,
 \widetilde{\mathbf{q}}^{N}].
\label{eq:semantic_queries}
\end{equation}
Unlike the image tokens, the semantic queries are never masked. The visible image tokens and semantic queries are concatenated and jointly processed by the shared SAREO Encoder $\mathcal{E}$ that yields three latents for SAR, EO and query:
\begin{equation}
[
\mathbf{Z}_{s}^{\mathrm{vis}};
\mathbf{Z}_{e}^{\mathrm{vis}};
\mathbf{Z}_{q}
]
=
\mathcal{E}
\left(
[
\mathbf{Y}_{s}^{\mathrm{vis}};
\mathbf{Y}_{e}^{\mathrm{vis}};
\widetilde{\mathbf{Q}}
]
\right).
\label{eq:encoder}
\end{equation}
where $[\cdot;\cdot;\cdot]$ indicates a concatenation operator. When a modality is unavailable or intentionally withheld, its token sequence is omitted from Equation~\ref{eq:encoder}. Through joint self-attention, the query latent $\mathbf{Z}_q$ aggregates spatially aligned evidence from all available modalities. Meanwhile, the SAR and EO latents, $\mathbf{Z}_{s}^{\mathrm{vis}}$ and $\mathbf{Z}_{e}^{\mathrm{vis}}$, exchange cross-modal context through both self-attention within concatenated three-modal streams.

\subsection{Decoupled Semantic Supervision}
\label{sec:semantic}

Masked reconstruction encourages image tokens to preserve low-level semantic representations such as local appearance, geometry, and sensing-specific characteristics. In contrast, a pretrained VFM operating on EO imagery encourages high-level semantic representations that are shared across sensing modalities. Directly applying both objectives to the same image tokens forces a single token stream to simultaneously satisfy these different representational demands. The SAREO-FM separates their explicit supervision destinations by applying the VFM target exclusively to the semantic-query stream. Given an EO image, a frozen VFM teacher $f_{\theta}$ processes the clean and unmasked observation, and extracts its final-layer patch tokens:
\begin{equation}
\mathbf{T}
=
f_{\theta}(\mathbf{X}_{e})
=
[\mathbf{t}^{1},\ldots,\mathbf{t}^{N}]
\in
\mathbb{R}^{N\times D_t},
\label{eq:teacher_tokens}
\end{equation}
where $\mathbf{t}^{i}$ denotes the teacher feature corresponding to the $i$-th spatial location, and $D_t$ is the teacher feature dimension. If the teacher produces a different spatial resolution, its patch-token grid is spatially resampled to match the $N$ query locations. A lightweight semantic projector
$h_{\phi}:\mathbb{R}^{D}\rightarrow\mathbb{R}^{D_t}$
maps the contextualized queries to the teacher feature space. We then impose patch-wise semantic alignment using cosine distance:
\begin{equation}
\mathcal{L}_{\mathrm{align}}
=
\frac{1}{N}
\sum_{i=1}^{N}
\left(
1-
\frac{
h_{\phi}(\mathbf{z}_{q}^{i})^{\top}\mathbf{t}^{i}
}{
\left\|h_{\phi}(\mathbf{z}_{q}^{i})\right\|_2
\left\|\mathbf{t}^{i}\right\|_2
}
\right).
\label{eq:alignment_loss}
\end{equation}

\noindent We refer to this design as \emph{decoupled semantic supervision}. The modality tokens are not directly constrained to reproduce EO-VFM features, and can therefore retain information useful for sensing-specific reconstruction. Instead, the dedicated query stream serves as the explicit destination of high-level semantic supervision. Importantly, this decoupling applies to the supervision heads rather than to the encoder computation: all three streams continue to exchange information, and gradients from both objectives jointly optimize the SAREO Encoder $\mathcal{E}$.

\subsection{Modality-Specific Masked Reconstruction}
\label{sec:decoder}

For pixel-level recovery, we employ a modality-specific reconstruction decoder $\mathcal{D}_m$ for $m\in\{s,e\}$. Consistent with our decoupled design, the semantic-query latents $\mathbf{Z}_q$ are discarded before decoding, and are not used as decoder conditions. For each modality, the encoded visible tokens are restored to their original spatial locations, and a shared learnable mask token $\mathbf{e}_m$ is inserted at every masked location. Specifically, we construct the ordered decoder input sequence
\begin{equation}
\mathbf{u}_m^{i}
=
\begin{cases}
\mathbf{z}_{m}^{\rho_m(i)},
& i\in\mathcal{V}_m,\\[2pt]
\mathbf{e}_m,
& i\in\Omega_m,
\end{cases}
\label{eq:token_restore}
\end{equation}
where $\rho_m(i)$ maps the original spatial index $i$ to its position in the packed visible-token sequence $\mathbf{Z}_{m}^{\mathrm{vis}}$. The decoder input at location $i$ is then
\begin{equation}
\mathbf{w}_m^{i}
=
\mathbf{u}_m^{i}
+
\mathbf{p}_{\mathrm{dec}}^{i}
+
\mathbf{a}_{m}^{\mathrm{dec}},
\quad
\mathbf{W}_m
=
[\mathbf{w}_m^1;\ldots;\mathbf{w}_m^N],
\label{eq:decoder_input}
\end{equation}
where $\mathbf{p}_{\mathrm{dec}}^{i}$ is a 2D sine--cosine decoder positional embedding and
$\mathbf{a}_{m}^{\mathrm{dec}}$ is a learnable decoder-specific modality embedding. The decoder predicts all image patches as:
\begin{equation}
[
\hat{\mathbf{x}}_{p,m}^{1};
\ldots;
\hat{\mathbf{x}}_{p,m}^{N}
]
=
\mathcal{D}_m(\mathbf{W}_m).
\label{eq:decoder_output}
\end{equation}
Following standard masked reconstruction, the pixel loss is evaluated only at masked locations:
\begin{equation}
\mathcal{L}_{\mathrm{pix}}^{m}
=
\frac{1}{|\Omega_m|}
\sum_{i\in\Omega_m}
\left\|
\hat{\mathbf{x}}_{p,m}^{i}
-
\mathbf{x}_{p,m}^{i}
\right\|_1.
\label{eq:pixel_loss}
\end{equation}
where, $\mathcal{L}_{\mathrm{pix}}^{s}$ and
$\mathcal{L}_{\mathrm{pix}}^{e}$ correspond to the SAR and EO pixel losses in Figure~\ref{fig:overview}, respectively.

\subsection{Modality Masking Strategy}
\label{sec:mixed_masking}

To expose the unified SAREO encoder $\mathcal{E}$ to diverse downstream input configurations, we train the SAREO-FM using four complementary masking modes:

\noindent\textbf{(i) Independent masking.}
SAR and EO masks are sampled independently. Visible patches from one modality can provide spatially aligned cues for locations masked in the other, encouraging fine-grained cross-modal reasoning.

\noindent\textbf{(ii) Shared masking.}
The same spatial mask is applied to both modalities. Because the corresponding location is simultaneously hidden in SAR and EO, the model must rely on broader spatial and semantic context rather than direct local correspondence.

\noindent\textbf{(iii) SAR-only input (EO masking).}
The EO token stream is entirely withheld from the encoder, while the SAR image is partially masked. The semantic queries must therefore extract semantic information solely from the available SAR observations.

\noindent\textbf{(iv) EO-only input (SAR masking).}
The SAR token stream is symmetrically withheld, leaving only partially masked EO tokens and the semantic queries as encoder inputs.

The first two modes, (i) and (ii), promote cross-modal interaction, whereas the modality-only modes, (iii) and (iv), ensure that the same encoder $\mathcal{E}$ remains effective when only one modality is available.

\subsection{Pretraining Objective}
\label{sec:objective}

Our pretraining objective combines modality-specific reconstruction and semantic alignment:
\begin{equation}
\mathcal{L}
=
\lambda_s\mathcal{L}_{\mathrm{pix}}^{s}
+
\lambda_e\mathcal{L}_{\mathrm{pix}}^{e}
+
\lambda_{\mathrm{align}}\mathcal{L}_{\mathrm{align}},
\label{eq:total}
\end{equation}
where each coefficient is set to zero when its supervision is unavailable or excluded by the sampled input mode. SAR--EO inputs activate four cases: (i) SAR-only inputs with EO masked use $\mathcal{L}_{\mathrm{pix}}^{s}$ and $\mathcal{L}_{\mathrm{align}}$, where EO is retained only for the frozen teacher; (ii) EO-only inputs with SAR masked use $\mathcal{L}_{\mathrm{pix}}^{e}$ and $\mathcal{L}_{\mathrm{align}}$; (iii) Unpaired SAR samples with EO absent contribute only to $\mathcal{L}_{\mathrm{pix}}^{s}$; and (iv) Unpaired EO samples also contribute to $\mathcal{L}_{\mathrm{pix}}^{e}$ and $\mathcal{L}_{\mathrm{align}}$.

\subsection{Downstream Transfer}
\label{sec:downstream}

During downstream transfer, the frozen VFM teacher $f_{\theta}$, semantic projector $h_{\phi}$, and reconstruction decoders $\{\mathcal{D}_m\}$ are discarded. The pretrained SAREO Encoder $\mathcal{E}$ receives all available, unmasked image tokens together with the semantic queries where a missing modality is represented by an empty token sequence, allowing $\mathcal{E}$ to process SAR, EO, or paired SAR--EO observations without modality-specific backbones or separate inference pipelines. $\mathcal{E}$ provides complementary representations for different downstream requirements. Modality-token outputs retain sensing-specific and spatially detailed information, whereas the semantic-query outputs provide contextually aggregated, semantically enriched features. Depending on the downstream tasks such as dense prediction or image-level recognition, their task-specific features can be selectively extracted from the modality tokens, the semantic queries, or their combination.
\section{Experiments}
\label{sec:experiments}

\subsection{Datasets and Downstream Tasks}
\label{sec:datasets}

\paragraph{Pretraining corpus.}
We pretrain SAREO-FM on SAR-1M~\cite{liu2026sarmae}, which aggregates 18 public SAR datasets spanning classification, detection, and segmentation scenarios. It contains 1,312,902 SAR images, of which 1,042,156 are paired with geographically aligned EO observations. Overall, the corpus covers 57 target and scene categories, multiple SAR sensors and polarization modes, and ground resolutions ranging from $0.1$ to $60$\,m.

\paragraph{Downstream benchmarks.}
Downstream heads use modality-token features ($\mathrm{mod}$), semantic-query features ($\mathrm{sem}$), or their concatenation ($\mathrm{both}$). We evaluate SAR recognition on MSTAR~\cite{ross1998mstar}, FUSAR-Ship~\cite{hou2020fusarship}, and SAR-ACD~\cite{sun2022scan}. MSTAR and FUSAR-Ship measure in-corpus transfer, as they are included in SAR-1M, whereas SAR-ACD serves as our primary out-of-corpus benchmark. EO-only transfer is evaluated on EuroSAT~\cite{helber2019eurosat}, NWPU-RESISC45~\cite{cheng2017resisc45}, and AID~\cite{xia2017aid} under few-shot and higher-supervision settings. For paired SAR--EO transfer, we evaluate single-label recognition on So2Sat LCZ42~\cite{zhu2020so2sat} and multi-label recognition on BigEarthNet-MM~\cite{sumbul2021bigearthnetmm}. We additionally examine qualitative dense transfer on BRIGHT~\cite{chen2025bright}, which requires building-damage assessment from paired pre-event EO and post-event SAR observations. We report the mean performance over three independent runs for end-to-end fine-tuning and five independent runs for linear probing.

\begin{table*}[t]
    \scriptsize
    \centering
    \caption{\textbf{SAR target classification with end-to-end fine-tuning.}
    The upper block reports published OA (\%) under the source-specific protocol
    of each method and is included only as context. The lower block reports our
    controlled evaluation with identical data splits and downstream pipelines;
    mean and standard deviation over three seeds are shown. $^*$ denotes our re-evaluation of the official SARMAE checkpoint. \textbf{Bold} denotes the best controlled result.}
    \label{tab:sar_finetune}
    \resizebox{1.0\textwidth}{!}{%
    \def\arraystretch{1.1}
    \setlength{\tabcolsep}{3pt}
    \begin{tabular}{@{}lll cc cc cc@{}}
        \toprule
        \multirow{2}{*}{\textbf{Method}}
        & \multirow{2}{*}{\textbf{Venue}}
        & \multirow{2}{*}{\textbf{Backbone}}
        & \multicolumn{2}{c}{\textbf{FUSAR-Ship}}
        & \multicolumn{2}{c}{\textbf{MSTAR}}
        & \multicolumn{2}{c}{\textbf{SAR-ACD}} \\
        \cmidrule(lr){4-5}
        \cmidrule(lr){6-7}
        \cmidrule(lr){8-9}
        & & &
        \textbf{40-shot} & \textbf{30\%}
        & \textbf{40-shot} & \textbf{30\%}
        & \textbf{40-shot} & \textbf{30\%} \\
        \midrule
        ResNet-50~\cite{resnet}
        & CVPR'16 & ResNet-50
        & -- & 58.41 & -- & 89.94 & -- & 59.70 \\
        Swin Transformer~\cite{swintransformer}
        & ICCV'21 & Swin-B
        & -- & 60.79 & -- & 82.97 & -- & 67.50 \\
        BEiT~\cite{beit}
        & ICLR'22 & ViT-B
        & 59.70 & 71.13 & 40.70 & 69.75 & -- & 79.77 \\
        CROMA~\cite{croma}
        & NeurIPS'23 & ViT-B
        & 83.71 & -- & -- & -- & -- & 88.99 \\
        SAR-JEPA~\cite{sarjepa}
        & ISPRS JPRS'24 & ViT-B
        & 85.80 & -- & 91.60 & -- & 75.50 & -- \\
        SARATR-X~\cite{saratrx}
        & TIP'25 & HiViT-B
        & 87.70 & -- & 98.10 & -- & 76.40 & -- \\
        LoMaR~\cite{lomar}
        & WACV'25 & ViT-B
        & 82.70 & -- & 77.00 & -- & 67.40 & -- \\
        SUMMIT~\cite{summit}
        & IJAEOG'25 & ViT-B
        & 81.50 & 71.91 & 63.60 & 98.39 & 68.70 & 84.25 \\
        Copernicus-FM~\cite{copernicusfm}
        & ICCV'25 & ViT-B
        & 87.61 & -- & -- & -- & -- & 92.63 \\
        CoDe-MAE~\cite{codemae}
        & arXiv'26 & HiViT-B
        & 89.40 & -- & 98.70 & -- & 78.30 & -- \\
        MaRS~\cite{mars}
        & AAAI'26 & SwinV2-B
        & 77.70 & -- & 75.50 & -- & 68.40 & -- \\
        SARMAE~\cite{liu2026sarmae}
        & CVPR'26 & ViT-B
        & 89.30 & 92.92 & 96.70 & 99.61 & -- & 95.06 \\
        SARMAE~\cite{liu2026sarmae}
        & CVPR'26 & ViT-L
        & 90.86 & 92.80 & 97.24 & 98.92 & -- & 95.63 \\
        \midrule
        Random initialization
        & -- & ViT-B
        & 60.31\sd{1.39} & 82.11\sd{0.52}
        & 34.71\sd{0.09} & 38.76\sd{8.35}
        & 33.14\sd{2.60} & 39.19\sd{10.27} \\
        SARMAE$^*$~\cite{liu2026sarmae}
        & CVPR'26 & ViT-B
        & 90.09\sd{0.36} & 92.82\sd{0.21}
        & 94.54\sd{1.72} & 97.18\sd{0.32}
        & 68.50\sd{2.60} & \textbf{95.01}\sd{0.85} \\
        \rowcolor{purple!5}
        \textbf{SAREO-FM ($\mathrm{sem}$)}
        & -- & ViT-B
        & 91.05\sd{0.16} & \textbf{93.13}\sd{0.14}
        & \textbf{97.57}\sd{0.42} & 99.13\sd{0.54}
        & \textbf{78.40}\sd{1.24} & 93.34\sd{0.55} \\
        \rowcolor{purple!5}
        \textbf{SAREO-FM ($\mathrm{both}$)}
        & -- & ViT-B
        & \textbf{91.41}\sd{0.23} & 93.12\sd{0.06}
        & 96.66\sd{0.31} & \textbf{99.36}\sd{0.47}
        & 78.00\sd{1.00} & 92.84\sd{0.68} \\
        \bottomrule
    \end{tabular}}
\end{table*}

\begin{figure}[t]
  \centering
  \begin{minipage}{0.49\textwidth}
  \centering
  \includegraphics[width=\textwidth]{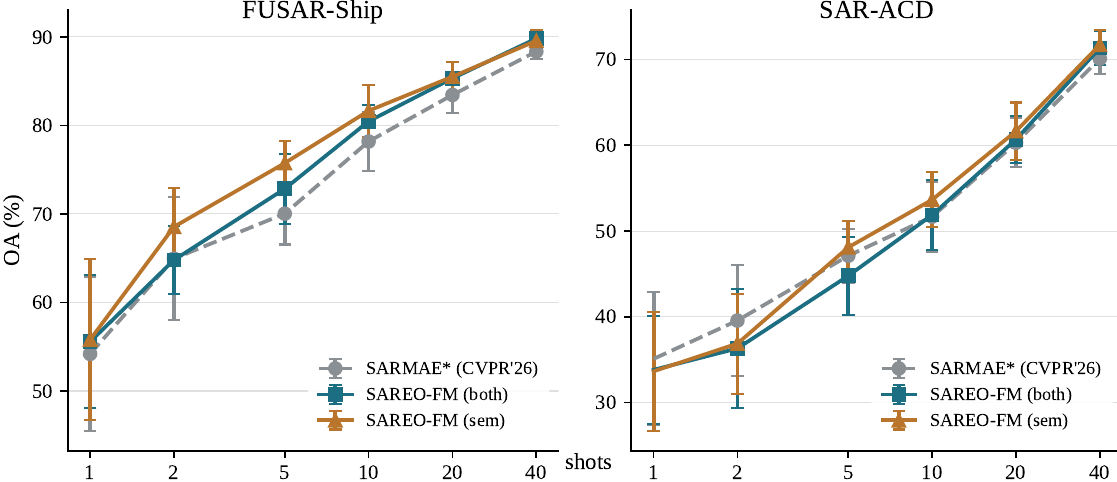}
  \caption{\textbf{Frozen few-shot SAR transfer.} We use fixed encoder features and train only the downstream classifier.}
  \label{fig:fewshot}
  \end{minipage}
  \hfill
  \begin{minipage}{0.48\textwidth}
  \includegraphics[width=\textwidth]{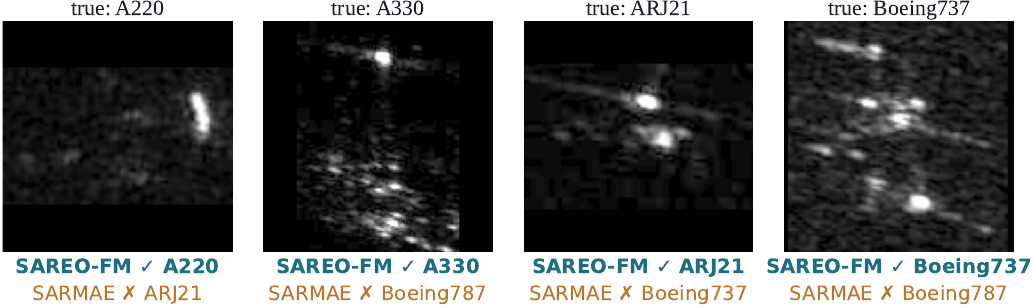}
  \caption{\textbf{Qualitative comparison on SAR-ACD.}
  Representative test images misclassified by SARMAE$^*$ but correctly
  classified by SAREO-FM.}
  \label{fig:saracd}
  \end{minipage}
\end{figure}

\subsection{Implementation Details}
\label{sec:implementation}

\paragraph{Pretraining.}
SAREO-FM uses a ViT-B/16 encoder~\cite{dosovitskiy2020image} trained from scratch on $256\times256$ inputs, yielding a $16\times16$ token grid. We accordingly introduce $256$ $(N=16\times16)$ learnable semantic queries. The teacher $f_\theta$ is a frozen DINOv3-7B model~\cite{simeoni2025dinov3}, and the semantic projector $h_\phi$ is a two-layer MLP. We empirically set the loss weights to $\lambda_{\mathrm{s}}=1$, $\lambda_{\mathrm{e}}=1$, and $\lambda_{\mathrm{align}}=0.5$. We sample independent masking, spatially shared masking, complete EO dropping, and complete SAR dropping with probabilities $0.35/0.35/0.15/0.15$. Each retained modality is masked by $75\%$, while all semantic queries remain active. We train for 300,000 steps using AdamW~\cite{loshchilov2019adamw} with $\beta=(0.9,0.95)$, weight decay $0.05$, and an effective batch size of 2,560. The learning rate reaches $1.5\times10^{-3}$ after 15,000 warm-up steps and then follows cosine decay to zero. Training uses bfloat16 (bf16) precision, gradient clipping at $1.0$, and four NVIDIA B200 GPUs, requiring approximately 390 B200 GPU-hours.

\paragraph{Controlled comparison.}
SARMAE~\cite{liu2026sarmae} serves as our primary controlled baseline. We denote by SARMAE$^*$ our reevaluation of its official pretrained ViT-B checkpoint under exactly the same downstream protocol as SAREO-FM, including dataset splits, sampled training subsets, input preprocessing, classifier heads, optimization settings, and evaluation code. Both methods employ ViT-B encoders pretrained on exactly the same SAR images from SAR-1M and leverage DINOv3 features~\cite{simeoni2025dinov3} extracted from corresponding EO observations. The key distinction is how the paired EO information is used: SARMAE uses EO-derived DINO features solely as semantic supervision for SAR representation learning, whereas our SAREO-FM additionally processes the raw EO observations as a second modality, and jointly learns SAR--EO reconstruction and cross-modal representations. SARMAE$^*$ therefore provides our closest controlled comparison, isolating the effect of incorporating raw paired EO observations and cross-modal learning.

\subsection{SAR Representation Transfer}
\label{sec:sar_transfer}

\paragraph{End-to-end fine-tuning.}
Table~\ref{tab:sar_finetune} first places our SAREO-FM among published SAR representation learners, and then provides a controlled comparison with SARMAE$^*$. The SAREO-FM obtains the best accuracy in five of the six controlled settings. On FUSAR-Ship, it improves over SARMAE$^*$ by $1.32$ points at 40 shots and $0.31$ points with 30\% supervision. The corresponding gains on MSTAR are $3.03$ and $2.18$ points. These results show that the benefit persists under substantially different supervision budgets, although both datasets overlap with the pretraining corpus. The strongest evidence comes from out-of-corpus SAR-ACD. With only 40 examples per class, semantic queries improve accuracy from $68.50\%$ to $78.40\%$, a $9.90$-point gain. Under 30\% supervision, however, SARMAE$^*$ remains $1.67$ points stronger. The contrast indicates that our SAREO-FM primarily improves label-efficient transfer rather than performance after abundant task-specific adaptation. Across the 40-shot columns, $\mathrm{sem}$ is the strongest on MSTAR and SAR-ACD, while $\mathrm{both}$ is the strongest on FUSAR-Ship, supporting task-dependent use of the two output streams.

\paragraph{Frozen few-shot transfer.}
Figure~\ref{fig:fewshot} removes encoder adaptation, and tests whether the pretrained representations are directly accessible to a lightweight classifier. On FUSAR-Ship, our SAREO-FM becomes consistently stronger than SARMAE$^*$ from 2-shot onward, with the largest separation occurring at 5-shot. On SAR-ACD, the advantage emerges at 5-shot and persists through 40-shot. The similar trajectories of $\mathrm{sem}$ and $\mathrm{both}$ show that the learned semantic queries remain useful even without end-to-end fine-tuning, while concatenating modality tokens can provide a small task-dependent complement. Figure~\ref{fig:saracd} complements the aggregate improvement with representative out-of-corpus examples. Our SAREO-FM correctly separates visually similar aircraft categories where the controlled SARMAE$^*$ baseline fails, consistent with the large 40-shot gain in Table~\ref{tab:sar_finetune}.

\subsection{EO Representation Transfer}
\label{sec:eo_transfer}

\begin{table*}[t]
    \scriptsize
    \centering
    \caption{\textbf{Frozen few-shot EO transfer.}
    We report accuracy (\%) for SAR and EO inputs over five support draws.
    CoDe-MAE values use its separate 10-shot protocol and are included only as
    context.}
    \label{tab:eo_transfer}
    \resizebox{\textwidth}{!}{%
    \def\arraystretch{1.1}
    \setlength{\tabcolsep}{3pt}
    \begin{tabular}{l c c c c c c c c}
        \toprule
        \multirow{2}{*}{\textbf{Method / Feature}}
        & \multicolumn{4}{c}{\textbf{SAR}}
        & \multicolumn{4}{c}{\textbf{EO}} \\
        \cmidrule(lr){2-5}
        \cmidrule(lr){6-9}
        & 1-shot & 5-shot
        & 10-shot & 40-shot
        & 1-shot & 5-shot
        & 10-shot & 40-shot \\
        \midrule
        DINOv3 (teacher)
        & 38.17\sd{4.61} & 50.72\sd{2.68}
        & 56.01\sd{1.81} & 64.35\sd{1.56}
        & 59.51\sd{3.53} & 75.96\sd{3.15}
        & 83.34\sd{1.42} & 90.88\sd{0.65} \\
        SARMAE$^*$
        & 37.22\sd{5.71} & 48.64\sd{3.00}
        & 53.97\sd{1.39} & 64.72\sd{1.12}
        & 43.56\sd{6.57} & 59.57\sd{3.05}
        & 66.49\sd{1.97} & 78.36\sd{1.07} \\
        CoDe-MAE
        & -- & -- & 59.88$^\dagger$ & --
        & -- & -- & 81.18$^\dagger$ & -- \\
        \rowcolor{purple!5}
        \textbf{SAREO-FM ($\mathrm{both}$)}
        & \textbf{39.07}\sd{5.98}
        & \textbf{53.54}\sd{2.21}
        & \textbf{59.36}\sd{1.46}
        & \textbf{69.24}\sd{0.96}
        & \textbf{64.41}\sd{5.05}
        & \textbf{81.09}\sd{1.61}
        & \textbf{86.76}\sd{1.24}
        & \textbf{92.58}\sd{0.75} \\
        \bottomrule
    \end{tabular}}
\end{table*}

Table~\ref{tab:eo_transfer} tests whether $\mathcal{E}$ transfers beyond SAR imagery. With EO input, our SAREO-FM consistently surpasses its frozen DINOv3 teacher, improving accuracy by $4.90$, $5.13$, $3.42$, and $1.70$ points from 1 to 40 shots. This result is notable because DINOv3 supplies the semantic target during pretraining: the SAREO-FM does not merely reproduce its features, but adapts them through joint SAR--EO context and reconstruction. The largest gains occur in the most label-scarce regimes, whereas the gap narrows as the number of labels increases. $\mathcal{E}$ also supports SAR-input EuroSAT recognition. Our SAREO-FM improves over the controlled SARMAE$^*$ baseline at every shot count, with gains ranging from $1.85$ to $5.39$ points, and reaches $69.24\%$ at 40 shots. The separately reported CoDe-MAE numbers provide protocol-specific context rather than a direct comparison. Together, Table~\ref{tab:eo_transfer} shows that the shared representation retains useful scene semantics from either modality rather than specializing exclusively in SAR target recognition.

\subsection{Joint SAR--EO Transfer}
\label{sec:joint_transfer}
\begin{wraptable}{r}{0.57\textwidth}
    \scriptsize
    \centering
    \vspace{-0.4cm}
    \caption{\textbf{Transfer on paired SAR--EO benchmarks.}
    We report OA (\%) on 5-shot So2Sat LCZ42 and micro-AP (\%) on
    5-shot BigEarthNet-MM.}
    \vspace{-0.2cm}
    \label{tab:multimodal}
    \resizebox{0.57\textwidth}{!}{%
    \def\arraystretch{1.1}
    \setlength{\tabcolsep}{3pt}
    \begin{tabular}{l l c c c}
        \toprule
        \multirow{2}{*}{\textbf{Input}} & \multirow{2}{*}{\textbf{Method}} & \multirow{2}{*}{\textbf{Feature}}
        & \multicolumn{2}{c}{\textbf{Benchmark}} \\
        \cmidrule(lr){4-5}
        & & & So2Sat LCZ42
        & BigEarthNet-MM \\
        \midrule
        SAR+EO
        & Random initialization
        & $\mathrm{mod}$
        & 40.56\sd{2.25}
        & 54.29\sd{2.30} \\
        EO
        & DINOv3 (teacher)
        & $\mathrm{mod}$
        & 54.25\sd{3.87}
        & \underline{64.73}\sd{1.43} \\
        \rowcolor{purple!5}
        EO
        & \textbf{SAREO-FM}
        & $\mathrm{mod}$
        & \underline{58.14}\sd{2.90}
        & 64.42\sd{1.44} \\
        SAR
        & SARMAE$^*$
        & $\mathrm{mod}$
        & 28.96\sd{1.68}
        & 57.44\sd{0.69} \\
        \rowcolor{purple!5}
        SAR
        & \textbf{SAREO-FM}
        & $\mathrm{mod}$
        & 29.59\sd{2.64}
        & 60.47\sd{0.79} \\
        \rowcolor{purple!5}
        SAR+EO
        & \textbf{SAREO-FM}
        & $\mathrm{both}$
        & \textbf{58.79}\sd{4.19}
        & \textbf{66.70}\sd{0.58} \\
        \bottomrule
    \end{tabular}}
    \vspace{-0.2cm}
\end{wraptable}
Table~\ref{tab:multimodal} evaluates whether paired observations provide information beyond either sensor alone. On BigEarthNet-MM, joint SAR--EO input reaches $66.70$ micro-AP, improving over the best SAREO-FM unimodal input by $2.28$ points and the DINOv3 EO baseline by $1.97$ points. The SAR branch itself improves over SARMAE$^*$ by $3.03$ points, while the EO branch performs similarly to DINOv3. The joint gain therefore reflects complementary SAR information rather than a uniformly stronger EO representation. On So2Sat LCZ42, the EO branch improves over DINOv3 by $3.89$ points, whereas adding SAR yields a smaller 0.65-point gain. Thus, multimodal fusion is beneficial when the downstream labels exploit complementary sensor cues, but is not guaranteed to improve every paired benchmark.

\begin{figure}
  \centering
  \includegraphics[width=0.75\textwidth]{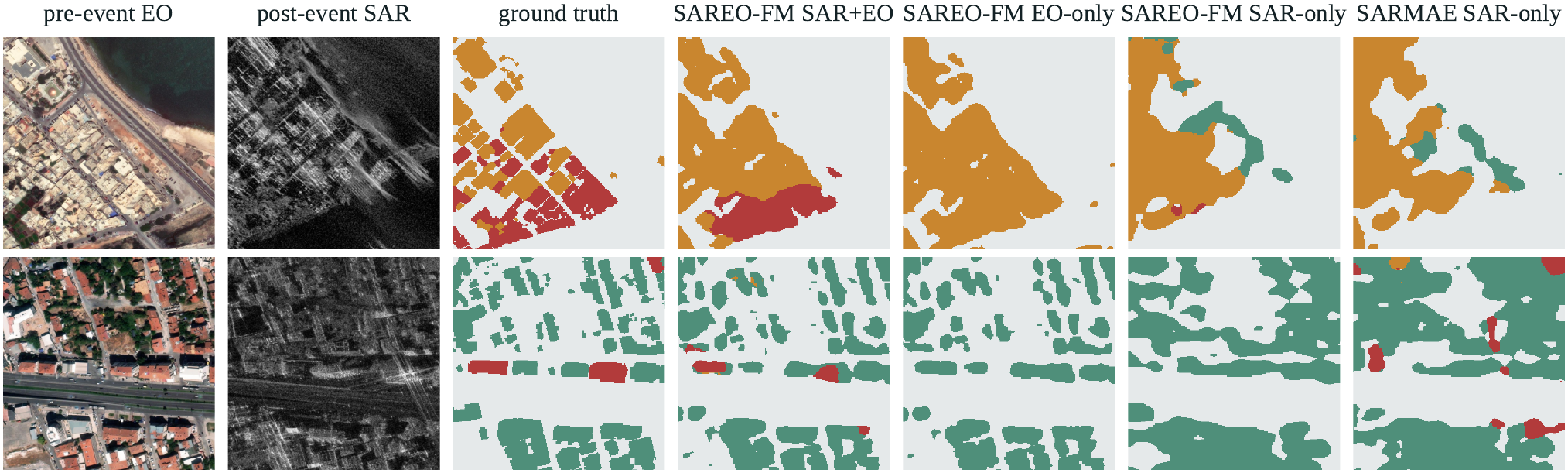}
  \caption{\textbf{Qualitative transfer on BRIGHT~\cite{chen2025bright}.}
  Given pre-event EO and post-event SAR observations, SAREO-FM fusion recovers
  destroyed buildings that the corresponding single-modality models predict
  as intact or background.}
  \vspace{-0.4cm}
  \label{fig:bright}
\end{figure}

\paragraph{Dense qualitative transfer.}
Although our quantitative evaluation focuses on image-level recognition, Figure~\ref{fig:bright} illustrates how the unified encoder can also support spatially aligned dense prediction. On BRIGHT, combining pre-event EO with post-event SAR recovers destroyed-building regions missed by EO-only and SAR-only inference, including the SARMAE$^*$ baseline. These examples are qualitative evidence rather than a substitute for a full segmentation benchmark, but they are consistent with the task-dependent fusion gain observed on BigEarthNet-MM.

\subsection{Ablation Study}
\label{sec:ablation}
\begin{wraptable}{r}{0.65\textwidth}
    \scriptsize
    \centering
    \vspace{-0.4cm}
    \caption{\textbf{Cumulative component ablation.}
    All variants use matched pretraining and evaluation protocols.
    We report 10-shot SAR linear-probing performance.
    Avg. denotes the arithmetic mean across the two benchmarks, while
    $\Delta$ measures the gain over the preceding configuration.}
    \vspace{-0.1cm}
    \label{tab:ablation}
    \resizebox{0.65\textwidth}{!}{%
    \def\arraystretch{1.1}
    \setlength{\tabcolsep}{3pt}
    \begin{tabular}{l c c c c}
        \toprule
        \textbf{Pretraining configuration}
        & \textbf{MSTAR (OA$\uparrow$)}
        & \textbf{FUSAR (OA$\uparrow$)}
        & \textbf{Avg. (OA$\uparrow$)}
        & \textbf{$\Delta$} \\
        \midrule
        Random initialization
        & 26.26\sd{1.86}
        & 47.86\sd{4.89}
        & 37.06
        & -- \\
        Plain MAE~\cite{he2022mae}
        & 66.16\sd{3.15}
        & 76.46\sd{2.52}
        & 71.31
        & $+34.25$ \\
        $+$ DINOv3 supervision
        & 73.49\sd{2.31}
        & 79.90\sd{2.49}
        & 76.70
        & $+5.39$ \\
        $+$ Decoupled semantic supervision
        & \underline{77.11}\sd{2.14}
        & \underline{82.55}\sd{2.15}
        & \underline{79.83}
        & $+3.13$ \\
        \rowcolor{purple!5}
        $+$ Raw EO modality (\textbf{SAREO-FM})
        & \textbf{78.41}\sd{2.14}
        & \textbf{84.00}\sd{2.88}
        & \textbf{81.21}
        & $+1.38$ \\
        \bottomrule
    \end{tabular}}
    \vspace{-0.1cm}
\end{wraptable}
We conduct a cumulative component ablation under matched pretraining and evaluation budgets. Each variant is pretrained from scratch for 30,000 steps with an effective batch size of 2,560. After pretraining, we freeze $\mathcal{E}$ and evaluate its SAR representation via 10-shot linear probing using the same 768-dimensional modality-feature readout. We report the mean and standard deviation over five seeds, using identical support sets for all variants. Table~\ref{tab:ablation} shows a consistent improvement as each component is introduced. Plain masked autoencoding establishes a strong representation-learning baseline, outperforming random initialization by 34.25 points on average. Adding supervision from the frozen DINOv3 teacher further improves the average by 5.39 points, indicating that semantic information derived from the geographically corresponding EO observations transfers effectively to SAR representations.

Decoupling semantic prediction from modality reconstruction yields a further 3.13-point gain. In particular, assigning semantic alignment to dedicated queries avoids requiring the modality tokens to simultaneously serve semantic prediction and pixel reconstruction, leading to more transferable SAR features. Finally, exposing the encoder to the raw paired EO observations and enabling extra EO reconstruction provides an additional 1.38-point improvement. Overall, the complete SAREO-FM improves upon plain MAE by 9.90 points on average, and achieves the best result on both benchmarks. These results show that teacher-based semantic supervision, objective decoupling, and raw cross-modal learning provide cumulative and complementary benefits.
\section{Conclusion}
\label{sec:conclusion}

We presented SAREO-FM, a unified foundation model for SAR, EO, and joint SAR--EO observations. The SAREO-FM assigns EO-VFM supervision to semantic queries and masked reconstruction to modality tokens, separating their explicit supervision while preserving their interaction within a shared encoder. The learned representations transfer effectively across SAR-only, EO-only, and multimodal downstream tasks, with joint observations providing benefits when the sensors offer complementary task-relevant information. Future work will extend this framework to additional sensing modalities and dense prediction tasks.

\clearpage
\appendix

\section*{Appendix}

\noindent
This \textit{Appendix} first anchors the additional analyses
to the principal results in the main paper, and then provides the
implementation and evaluation details needed to interpret them.
Section~\ref{sec:add_results} reports the exact frozen few-shot SAR results and
defines the scope of the controlled \sarmaestar{}~\cite{liu2026sarmae} comparison.
Section~\ref{sec:masking} makes the four masking modes, executed sampling
probabilities, encoder sequence lengths, and active losses explicit.
Section~\ref{sec:data} documents corpus membership, downstream splits, and the
interpretation of in-corpus and out-of-corpus transfer.
Section~\ref{sec:implementation_sup} records the architecture actually used by the
reported checkpoint. Section~\ref{sec:downstream_sup} defines the feature readouts
and presents checkpoint-only readout and fusion controls. Finally,
Sections~\ref{sec:efficiency} and \ref{sec:limitations} discuss deployment cost,
reproducibility, and the appropriate scope of the empirical claims.

\begin{table}[h]
    \scriptsize
    \centering
    \caption{Overview of the \textit{Appendix}.}
    \label{tab:supple_overview}
    \resizebox{0.55\textwidth}{!}{%
    \def\arraystretch{1.1}
    \setlength{\tabcolsep}{3pt}
    \begin{tabular}{cl}
        \toprule
        \rowcolor{purple!5}
        \textbf{Section} & \textbf{Contents} \\
        \midrule
        Section~\ref{sec:add_results} & Main-paper results and additional controls \\
        Section~\ref{sec:masking} & Masking modes and training objectives \\
        Section~\ref{sec:data} & Pre-training corpus, datasets, and evaluation scope \\
        Section~\ref{sec:implementation_sup} & Architecture and pre-training details \\
        Section~\ref{sec:downstream_sup} & Feature readouts and downstream protocols \\
        Section~\ref{sec:efficiency} & Inference cost \\
        Section~\ref{sec:limitations} & Limitations \\
        \bottomrule
    \end{tabular}}
\end{table}

\section{Additional Results and Controls}
\label{sec:add_results}

\subsection{Exact Frozen Few-Shot SAR Results}
\label{sec:exact_sar}

\begin{table*}[t]
    \scriptsize
    \centering
    \caption{\textbf{Exact frozen few-shot SAR classification results
    corresponding to Fig.~3 of the main paper.}
    We freeze the pretrained encoder and optimize only a linear downstream
    classifier. We report mean OA (\%) and sample standard deviation over five
    matched support draws. $\mathrm{sem}$ uses semantic-query features, whereas
    $\mathrm{both}$ concatenates modality-token and semantic-query features.
    \sarmaestar{} denotes the official checkpoint evaluated with the same
    downstream protocol. \textbf{Bold} and \underline{underline} denote the best and second-best
    values within each dataset and shot count, respectively; shaded rows are
    ours.}
    \label{tab:sar_fewshot}
    \resizebox{1.0\textwidth}{!}{%
    \def\arraystretch{1.1}
    \setlength{\tabcolsep}{3pt}
    \begin{tabular}{l l c c c c c c}
        \toprule
        \textbf{Dataset} & \textbf{Method / Feature}
        & \textbf{1-shot} & \textbf{2-shot} & \textbf{5-shot}
        & \textbf{10-shot} & \textbf{20-shot} & \textbf{40-shot} \\
        \midrule
        & \sarmaestar
        & 54.19\sd{8.69} & \underline{64.95}\sd{6.93} & 70.03\sd{3.48}
        & 78.18\sd{3.28} & 83.44\sd{2.03} & 88.35\sd{0.83} \\
        \rowcolor{purple!5}\cellcolor{white}\multirow{-1}{*}{FUSAR-Ship}
        & \textbf{SAREO-FM ($\mathrm{both}$)}
        & \underline{55.56}\sd{7.50} & 64.79\sd{3.81} & \underline{72.84}\sd{3.97}
        & \underline{80.46}\sd{1.81} & \underline{85.32}\sd{1.79}
        & \textbf{89.81}\sd{0.70} \\
        \rowcolor{purple!5}\cellcolor{white}
        & \textbf{SAREO-FM ($\mathrm{sem}$)}
        & \textbf{55.79}\sd{9.07} & \textbf{68.54}\sd{4.35}
        & \textbf{75.74}\sd{2.46} & \textbf{81.65}\sd{2.91}
        & \textbf{85.50}\sd{1.71} & \underline{89.56}\sd{1.22} \\
        \midrule
        & \sarmaestar
        & \textbf{35.09}\sd{7.77} & \textbf{39.56}\sd{6.52}
        & \underline{47.12}\sd{3.16} & 51.68\sd{4.08}
        & 60.30\sd{2.86} & 70.11\sd{1.82} \\
        \rowcolor{purple!5}\cellcolor{white}
        & \textbf{SAREO-FM ($\mathrm{both}$)}
        & \underline{33.81}\sd{6.30} & 36.31\sd{6.91}
        & 44.78\sd{4.56} & \underline{51.87}\sd{4.12}
        & \underline{60.69}\sd{2.75} & \underline{71.35}\sd{2.01} \\
        \rowcolor{purple!5}\cellcolor{white}\multirow{-3}{*}{SAR-ACD}
        & \textbf{SAREO-FM ($\mathrm{sem}$)}
        & 33.64\sd{6.95} & \underline{36.87}\sd{5.83}
        & \textbf{48.10}\sd{3.02} & \textbf{53.65}\sd{3.19}
        & \textbf{61.65}\sd{3.35} & \textbf{71.72}\sd{1.74} \\
        \bottomrule
    \end{tabular}}
\end{table*}

Table~\ref{tab:sar_fewshot} provides the numerical values underlying the
frozen-transfer trends in the main paper. On the in-corpus FUSAR-Ship
benchmark, the strongest \method{} readout is numerically above
\sarmaestar{} at every evaluated supervision budget, with gains of
$1.60$, $3.59$, $5.71$, $3.47$, $2.06$, and $1.46$ points from 1 to
40 shots. The largest
separation occurs at 5 shots. On the out-of-corpus SAR-ACD benchmark,
\sarmaestar{} remains stronger in the extremely sparse 1- and 2-shot
settings. The semantic-query readout crosses over at 5 shots and improves
the baseline by $0.98$, $1.97$, $1.35$, and $1.61$ points at 5, 10, 20,
and 40 shots, respectively. Across the twelve dataset--shot combinations,
the semantic-query readout is the stronger of the two \method{} readouts
in ten cases. This pattern indicates that semantic queries usually provide
the most direct frozen representation for classification, while
modality-token features offer a task-dependent complement. These are
numerical comparisons; we do not infer statistical significance from
Table~\ref{tab:sar_fewshot} alone.

\section{Masking and Objectives}
\label{sec:masking}

\subsection{Encoder Sequence Composition}
\label{sec:token_counts}

At $256\times256$ resolution, each modality is represented by a
$16\times16$ grid of 256 image tokens. A 75\% mask leaves 64 visible
tokens for each retained modality. All 256 semantic queries remain active
in every training and inference configuration. Table~\ref{tab:token_counts}
reports the resulting encoder sequence lengths. The token counts exclude
the frozen teacher and the decoder-side mask tokens, neither of which is
part of the shared encoder sequence.

\begin{table}[t]
    \scriptsize
    \centering
    \caption{\textbf{Shared-encoder sequence composition.}
    Training counts use the 75\% patch-mask ratio; downstream inference
    uses all available image patches.}
    \label{tab:token_counts}
    \resizebox{0.5\columnwidth}{!}{%
    \def\arraystretch{1.1}
    \setlength{\tabcolsep}{3pt}
    \begin{tabular}{lcccc}
        \toprule
        \rowcolor{gray!5}\textbf{Configuration} & \textbf{SAR} & \textbf{EO}
        & \textbf{Queries} & \textbf{Total} \\
        \midrule
        Independent/shared masking & 64 & 64 & 256 & 384 \\
        Complete EO dropping & 64 & 0 & 256 & 320 \\
        Complete SAR dropping & 0 & 64 & 256 & 320 \\
        Unpaired SAR training & 64 & 0 & 256 & 320 \\
        \midrule
        SAR-only inference & 256 & 0 & 256 & 512 \\
        EO-only inference & 0 & 256 & 256 & 512 \\
        Joint SAR--EO inference & 256 & 256 & 256 & 768 \\
        \bottomrule
    \end{tabular}}
\end{table}

\subsection{Masking Modes and Active Losses}
\label{sec:mask_loss}

Table~\ref{tab:mask_loss} distinguishes patch masking from complete
modality dropping. ``Dropped'' means that the corresponding image-token
stream is omitted entirely from the student encoder. It does not mean
that the model is trained to reconstruct a completely absent modality.
When EO is dropped from a paired student input, the clean EO observation
remains available only to the frozen teacher and supplies semantic
targets for the query stream.

\begin{table*}[t]
    \scriptsize
    \centering
    \caption{\textbf{Student inputs, teacher availability, and active
    losses.} Retained image streams are masked by 75\%, and all semantic
    queries remain active. The loss coefficients are
    $\lambda_s=1$, $\lambda_e=1$, and
    $\lambda_{\mathrm{align}}=0.5$.}
    \label{tab:mask_loss}
    \resizebox{0.9\textwidth}{!}{%
    \def\arraystretch{1.1}
    \setlength{\tabcolsep}{3pt}
    \begin{tabular}{lclll}
        \toprule
        \rowcolor{gray!5}\textbf{Sample / mode} & \textbf{Prob.}
        & \textbf{Student encoder image streams}
        & \textbf{Teacher input} & \textbf{Active objective} \\
        \midrule
        Paired, independent masks
        & 0.35 & Visible SAR + visible EO & Clean EO
        & $\mathcal{L}_{\mathrm{pix}}^s+
           \mathcal{L}_{\mathrm{pix}}^e+
           0.5\mathcal{L}_{\mathrm{align}}$ \\
        Paired, shared mask
        & 0.35 & Visible SAR + visible EO & Clean EO
        & $\mathcal{L}_{\mathrm{pix}}^s+
           \mathcal{L}_{\mathrm{pix}}^e+
           0.5\mathcal{L}_{\mathrm{align}}$ \\
        Paired, complete EO dropping
        & 0.15 & Visible SAR only & Clean EO, teacher only
        & $\mathcal{L}_{\mathrm{pix}}^s+
           0.5\mathcal{L}_{\mathrm{align}}$ \\
        Paired, complete SAR dropping
        & 0.15 & Visible EO only & Clean EO
        & $\mathcal{L}_{\mathrm{pix}}^e+
           0.5\mathcal{L}_{\mathrm{align}}$ \\
        \midrule
        Unpaired SAR
        & -- & Visible SAR only & Unavailable
        & $\mathcal{L}_{\mathrm{pix}}^s$ \\
        Unpaired EO
        & -- & Not sampled in this work & -- & -- \\
        \bottomrule
    \end{tabular}}
\end{table*}

The values in Table~\ref{tab:mask_loss} are the code- and log-verified
sampling probabilities used for the reported checkpoint. Across 357,422
recorded batches, the realized frequencies were
$0.3506/0.3506/0.1493/0.1494$, closely matching the configured
$0.35/0.35/0.15/0.15$ distribution.

The general formulation can accommodate an unpaired EO corpus, but the
SAR-1M~\cite{liu2026sarmae} run used in this work consists of paired SAR--EO observations and
unpaired SAR observations. We therefore do not present unpaired EO as a
sample type instantiated in the reported training run. Likewise, no pixel
loss is applied to an image stream that is completely absent from the
student encoder.

\subsection{One Pretraining Iteration}
\label{sec:training_step}

For completeness, one optimization step can be summarized as follows.

\begin{enumerate}
    \item Draw either a paired SAR--EO record or an unpaired SAR record
    from the pretraining corpus.
    \item For a paired record, sample one of the four modes in
    Table~\ref{tab:mask_loss}; for every retained modality, sample an
    exact 75\% patch mask.
    \item Concatenate the visible image tokens and all 256 semantic
    queries, and process them with the shared encoder.
    \item For each retained target modality, restore its visible outputs to
    the spatial grid, insert decoder mask tokens, and predict its image
    patches with the corresponding modality-specific decoder.
    \item If a paired EO observation is available, process its clean,
    unmasked teacher view with the frozen DINOv3 model, resample the
    teacher patch grid to $16\times16$ when needed, and align it
    patch-wise with the projected query outputs.
    \item Sum only the losses activated by Table~\ref{tab:mask_loss} and
    update the shared encoder, queries, patch embeddings, decoders, and
    semantic projector. The DINOv3~\cite{simeoni2025dinov3} teacher remains frozen.
\end{enumerate}

\section{Data and Evaluation Scope}
\label{sec:data}

\subsection{Pretraining Corpus Composition}
\label{sec:corpus}

\method{} is pretrained on SAR-1M~\cite{liu2026sarmae}, which contains 1,312,902 SAR
observations aggregated from 18 public datasets. Among them, 1,042,156
have geographically corresponding EO observations and the remaining
270,746 are unpaired SAR observations. The corpus covers 57 target and
scene categories, multiple SAR sensors and polarization modes, and ground
resolutions from 0.1 to 60~m. Paired EO is used in two distinct ways:
as a raw student input when the sampled mode retains EO, and as the clean
input to the frozen semantic teacher. Unpaired SAR samples contribute only
SAR masked reconstruction.

\subsection{Downstream Benchmark Summary}
\label{sec:dataset_summary}

Table~\ref{tab:dataset_summary} summarizes the role of each downstream
benchmark. ``In corpus'' describes image-source membership in SAR-1M, not
supervised label leakage: no downstream labels are used during
pretraining. We reserve the term ``out of corpus'' for benchmarks whose
source observations are absent from SAR-1M. Benchmarks for which we make
no out-of-corpus claim are reported simply as downstream transfer tasks.

\begin{table*}[t]
    \scriptsize
    \centering
    \caption{\textbf{Downstream datasets and interpretation.}
    OA denotes overall accuracy; micro-AP denotes micro-averaged average
    precision. The paired EuroSAT protocol uses geospatially matched SAR
    observations in addition to standard EO imagery; standard EuroSAT
    itself is EO-only.}
    \label{tab:dataset_summary}
    \resizebox{\textwidth}{!}{%
    \def\arraystretch{1.1}
    \setlength{\tabcolsep}{3pt}
    \begin{tabular}{lccclll}
        \toprule
        \rowcolor{gray!5}\textbf{Dataset} & \textbf{Input} & \textbf{Classes}
        & \textbf{Metric} & \textbf{Protocol}
        & \textbf{SAR-1M relation} & \textbf{Role in this work} \\
        \midrule
        MSTAR & SAR & 10 & OA
        & 1--40-shot / 30\% & In corpus & In-corpus target transfer \\
        FUSAR-Ship & SAR & 10 & OA
        & 1--40-shot / 30\% & In corpus & In-corpus ship transfer \\
        SAR-ACD & SAR & 5 & OA
        & 1--40-shot / 30\% & Out of corpus & Primary out-of-corpus SAR test \\
        \midrule
        EuroSAT protocol & EO or SAR & 10 & OA
        & 1/5/10/40-shot & No out-of-corpus claim & Unimodal scene transfer \\
        NWPU-RESISC45 & EO & 45 & OA
        & Few-shot / higher supervision & No out-of-corpus claim & EO scene transfer \\
        AID & EO & 30 & OA
        & Few-shot / higher supervision & No out-of-corpus claim & EO scene transfer \\
        \midrule
        So2Sat LCZ42 & SAR, EO, or joint & 17 & OA
        & 5-shot, city-disjoint & No out-of-corpus claim & Paired single-label transfer \\
        BigEarthNet-MM & SAR, EO, or joint & 19 & micro-AP
        & 5-shot, official split & No out-of-corpus claim & Paired multi-label transfer \\
        BRIGHT & Pre-EO + post-SAR & 4 & Dense labels
        & Qualitative only & In-corpus source & In-corpus dense diagnostic \\
        \bottomrule
    \end{tabular}}
\end{table*}

\subsection{Corpus Membership and Geographic Overlap}
\label{sec:overlap}

MSTAR and FUSAR-Ship are constituents of SAR-1M and are explicitly
interpreted as in-corpus representation transfer. SAR-ACD is absent from
SAR-1M and provides the primary out-of-corpus SAR classification result.
BRIGHT also contributes source imagery to SAR-1M; consequently,
Figure~\ref{fig:bright} of the main paper is an in-corpus qualitative dense probe rather
than evidence of unseen-event or unseen-scene generalization. Its purpose
is to illustrate that the unified encoder can be attached to a dense
prediction head and can use complementary pre-event EO and post-event SAR
evidence.

For paired remote-sensing corpora, exact duplicate detection alone is not
sufficient to characterize overlap because resized crops can differ at
the byte level while originating from the same geographic tile, parent
scene, or event. A complete audit therefore uses source identifiers,
geographic coordinates, parent-scene identifiers, and event identifiers
in addition to file hashes. We do not use EuroSAT, So2Sat LCZ42,
BigEarthNet-MM, or BRIGHT to support an out-of-corpus claim unless such a
source-level exclusion is established.

\input{generated-sections/overlap-audit.tex}

\subsection{Paired-Dataset Splits}
\label{sec:paired_splits}

So2Sat LCZ42 is evaluated using a city-disjoint split so that geographic
regions do not cross the downstream train and test partitions. For each
few-shot draw, the same class-balanced support identifiers are used by all
compared input modes and feature readouts. BigEarthNet-MM contains 519,284
paired observations in the protocol used here, partitioned into
269,695/123,723/125,866 train/validation/test samples, respectively. We
map the labels to 19 classes and report micro-AP. Support sampling is
performed only within the training partition; validation is used for
downstream model selection, and the test partition remains fixed.

The ``SAR input'' rows associated with EuroSAT use a geospatially paired
SAR companion rather than the EO-only EuroSAT release by itself. The
accompanying dataset manifest records the companion release, VV/VH channel
ordering, geographic matching key, and paired-location split. These
identifiers are treated as part of the evaluation protocol rather than as
an implementation detail.

\input{generated-sections/preprocessing.tex}

\section{Architecture and Pretraining}
\label{sec:implementation_sup}

\subsection{Architecture Specification}
\label{sec:architecture}

The shared encoder follows ViT-B/16~\cite{dosovitskiy2020image}: 12 Transformer blocks, hidden
dimension $D=768$, 12 attention heads, and an multi-layer perceptron (MLP) expansion ratio of 4.
SAR and EO use separate linear patch projections but share the same
$16\times16$ two-dimensional sine--cosine positional grid. Learnable
stream embeddings distinguish SAR tokens, EO tokens, and semantic
queries. A spatially organized bank of 256 learnable queries uses the
same positional grid as the image tokens, establishing one query per
patch location.

The frozen DINOv3-7B~\cite{simeoni2025dinov3} teacher receives a clean EO view and supplies
final-layer patch features. If its native token grid differs from
$16\times16$, the patch features are spatially resampled to the query
grid before patch-wise cosine alignment. A two-layer MLP projects each
query output from the encoder dimension to the teacher-feature
dimension. SAR and EO have separate reconstruction decoders and mask
tokens.

\begin{table}[t]
    \scriptsize
    \centering
    \caption{\textbf{Architecture summary.}}
    \label{tab:architecture}
    \resizebox{0.55\textwidth}{!}{%
    \def\arraystretch{1.1}
    \setlength{\tabcolsep}{3pt}
    \begin{tabular}{ll}
        \toprule
        \rowcolor{gray!5}\textbf{Component} & \textbf{Specification} \\
        \midrule
        Input resolution & $256\times256$ \\
        Patch size / grid & $16\times16$ / $16\times16$ \\
        SAR / EO tokens & 256 per available modality \\
        Shared encoder & ViT-B/16, 12 blocks, $D=768$ \\
        Attention / MLP & 12 heads / ratio 4 \\
        Semantic queries & 256 learnable, spatially indexed \\
        Position encoding & Shared 2D sine--cosine grid \\
        Stream encoding & Learnable SAR / EO / query embeddings \\
        Teacher & Frozen DINOv3-7B patch features \\
        Semantic projector & Two-layer MLP \\
        Reconstruction & Separate SAR/EO decoders \\
        \bottomrule
    \end{tabular}}
\end{table}

\input{generated-sections/architecture-code-details.tex}

\subsection{Component Lifecycle}
\label{sec:lifecycle}

Table~\ref{tab:lifecycle} separates pretraining-only modules from the
deployed foundation model. The DINOv3 teacher, semantic projector, and
both reconstruction decoders are removed after pretraining. Therefore,
the large teacher contributes neither parameters nor computation to
downstream inference.

\begin{table}[t]
    \scriptsize
    \centering
    \caption{\textbf{Lifecycle of \method{} components.}}
    \label{tab:lifecycle}
    \resizebox{0.6\textwidth}{!}{%
    \def\arraystretch{1.1}
    \setlength{\tabcolsep}{3pt}
    \begin{tabular}{lcc}
        \toprule
        \rowcolor{gray!5}\textbf{Component} & \textbf{Pretraining} & \textbf{Downstream} \\
        \midrule
        Shared SAREO encoder & Trainable & Used \\
        Semantic queries & Trainable & Used \\
        SAR/EO patch embeddings & Trainable & Used as available \\
        SAR/EO reconstruction decoders & Trainable & Discarded \\
        Semantic projector & Trainable & Discarded \\
        DINOv3-7B teacher & Frozen & Discarded \\
        \bottomrule
    \end{tabular}}
\end{table}

\subsection{Pixel and Semantic Targets}
\label{sec:targets}

For each retained modality, the reconstruction decoder predicts all 256
patches, but the $\ell_1$ pixel objective is evaluated only on the
masked locations. The SAR and EO losses are not compared numerically
across modalities because their target spaces and normalization differ.
For semantic alignment, the projected query at each spatial location is
matched to the corresponding teacher patch by cosine distance. All 256
queries participate in semantic alignment, including locations whose SAR
or EO image patches are hidden from the student. This design encourages
the query grid to infer spatial semantics from the visible evidence and
context rather than copy a teacher feature from an identically visible
patch.

\section{Downstream Protocol}
\label{sec:downstream_sup}

\subsection{Feature Readouts}
\label{sec:readout}

During downstream transfer, all available image patches and all semantic
queries are passed to the shared encoder without masking. Let
$\mathcal{A}$ denote the set of available modalities and let
$Z_m=\{z_m^i\}_{i=1}^{N}$ and
$Z_q=\{z_q^i\}_{i=1}^{N}$ be the corresponding encoder outputs.
We first pool within each modality and then concatenate the available
modality streams:
\begin{align}
  \bar z_m
  &= \frac{1}{N}\sum_{i=1}^{N}z_m^i,
  \\
  g_{\mathrm{mod}}
  &= \underset{m\in\mathcal{A}}{\operatorname{concat}}\,\bar z_m,
  \\
  g_{\mathrm{sem}}
  &= \frac{1}{N}\sum_{i=1}^{N}z_q^i,
  \\
  g_{\mathrm{both}}
  &= [g_{\mathrm{mod}};g_{\mathrm{sem}}].
\end{align}
Table~\ref{tab:readout_dims} gives the resulting dimensions. In particular,
$\mathrm{mod}$ is not an average across modalities: it is 768-D for unimodal
input and 1,536-D for paired input. The semantic readout remains 768-D,
making $\mathrm{both}$ 1,536-D or 2,304-D, respectively.

\begin{table}[t]
    \scriptsize
    \centering
    \caption{\textbf{Image-level readout dimensions.}}
    \label{tab:readout_dims}
    \resizebox{0.35\textwidth}{!}{%
    \def\arraystretch{1.1}
    \setlength{\tabcolsep}{3pt}
    \begin{tabular}{lccc}
        \toprule
        \rowcolor{gray!5}\textbf{Available input} & $\mathrm{mod}$ & $\mathrm{sem}$
        & $\mathrm{both}$ \\
        \midrule
        SAR or EO & 768 & 768 & 1,536 \\
        SAR + EO & 1,536 & 768 & 2,304 \\
        \bottomrule
    \end{tabular}}
\end{table}

For $K$ classes, the corresponding linear heads contain
$K(768|\mathcal{A}|+1)$, $K(768+1)$, and
$K(768(|\mathcal{A}|+1)+1)$ parameters for $\mathrm{mod}$, $\mathrm{sem}$, and
$\mathrm{both}$, respectively. We therefore treat $\mathrm{both}$ as a
complementary representation and do not attribute every difference to
feature content alone.

\section{Inference Cost}
\label{sec:efficiency}

The deployed model contains the shared encoder, the two patch embeddings,
the persistent semantic queries, and the selected downstream head.
Teacher features are never computed at transfer time, and neither the
semantic projector nor the reconstruction decoders are loaded for normal
downstream inference. The exact shared-encoder sequence length is 512
tokens for either unimodal input and 768 tokens for joint input
(Table~\ref{tab:token_counts}). These sequence lengths explain why
unimodal and joint inference should be profiled separately.

Measured latency, throughput, peak memory, and FLOPs depend on hardware,
precision, batch size, and kernel implementation. We therefore report
them only under one explicitly fixed profiling protocol rather than
extrapolating from token counts. The checkpoint-only workflow uses
$256\times256$ inputs, evaluation mode, inference-only execution, fixed
warm-up and measurement iterations, explicit GPU synchronization, and
identical precision across modes. It reports the deployed encoder
separately from pretraining-only parameters.

\section{Limitations}
\label{sec:limitations}

Our quantitative evaluation focuses on image-level classification and
multi-label recognition. The BRIGHT examples demonstrate compatibility
with a dense head but are qualitative, in-corpus diagnostics and should
not be interpreted as a complete segmentation benchmark. The model also
assumes that paired SAR and EO observations are sufficiently registered
for patch-wise interaction and teacher alignment. Geographic
misregistration, acquisition-time differences, clouds, SAR layover, and
sensor-dependent resolution can weaken local correspondence. Finally,
the semantic target is obtained from an EO-pretrained VFM; although SAR
reconstruction preserves sensing-specific information, semantics that are
poorly represented by the EO teacher may be transferred less effectively.

\clearpage

\bibliography{iclr2027_conference}
\bibliographystyle{iclr2027_conference}

\end{document}